# Robust Dead Reckoning: Calibration, Covariance Estimation, Fusion and Integrity Monitoring

Maximilian Harr, Christoph Schaefer
Opel Automobile GmbH, EE Advanced Technology, Active Safety & Controls, Bahnhofsplatz, 65423 Russelsheim am Main
Email: {maximilian.harr, christoph.schaefer}@opel.de

**Abstract**
To measure system states and local environment directly with high precision, expensive sensors are required. However, highly accurate system states and environmental perception can also be achieved using data fusion techniques and digital maps. One crucial task of multi-sensor state estimation is to project different sensor measurements into the same temporal, spatial and physical domain, estimate their covariance matrices as well as the exclusion of erroneous measurements.
This paper presents a generic approach for robust estimation of vehicle movement (odometry). We will shortly present our calibration procedure, including the estimation of sensor alignments, offset / scaling errors, covariances / correlations and time delays. An improved algorithm for wheel diameter estimation is presented. Additionally an approach for robust odometry will be shown as odometry estimations are fused under known covariances, while outliers are detected using a chi-squared test. Utilizing our robust odometry, local environmental views can be associated and fused. Furthermore our robust odometry can be used to detect and exclude erroneous position estimates.

## I. Introduction and Related Work

### A. Introduction

Precise knowledge of vehicle system states and local environment is an essential task for Intelligent Transportation Systems (ITS) especially for Advanced Driver Assistance Systems (ADAS[1]). One crucial task of enhanced ADAS systems is the collection of high-precision digital map data. The BMWi-funded project Ko-HAF [5] addresses this challenge towards fully autonomous driving vehicles. The goal is to exchange digital map data and dynamic content among a broad range of automotive companies to ensure an enlarged field of view to make driving on highways safer and more efficient. A safety server processes, fuses and filters the environmental perception of all vehicles and provides a collective perception to participating vehicles.

[1]e.g. Adaptive Cruise Control (ACC), Automatic Parking, Forward Collision Warning, Emergency Driver Assistant and particularly Automated Driving.

The idea is to locally fuse environmental perceptions of multiple sensor measurements, such as detected lanes, traffic signs and hazardous obstacles, into one single perception, that will be send to the server to decrease data traffic. Fig. 1 shows the local mapping process. A precise spatial transformation consisting of translation $\vec{t}$ and rotation $R$ between each environmental snapshot (commonly referred to as dead reckoning) is thus required. Note that only GNSS position measurements will be used. Landmark-based localization techniques are not used to prevent coupling of erroneous landmarks from digital maps into the newly detected ones. Consequentially high precision requirements follow for transformation with increasing size of local views.

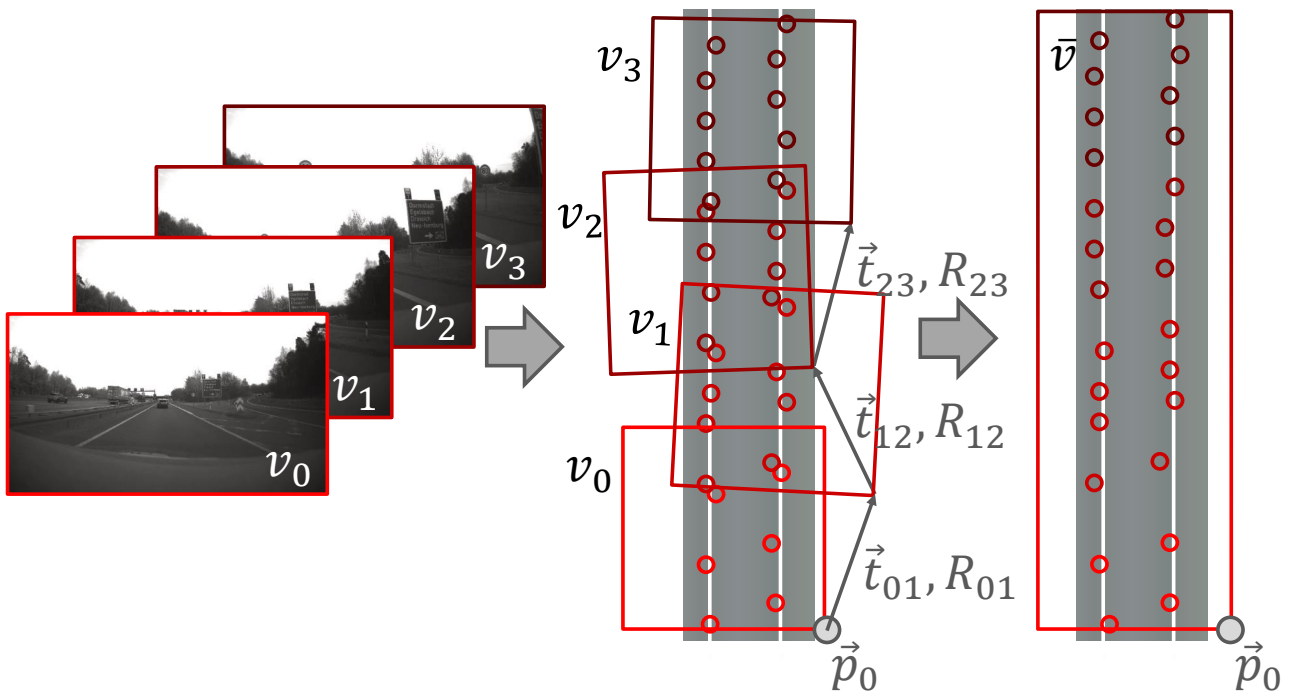

Fig. 1. **Local mapping process.** Multiple local environment snapshots need to be fused in a single local view.

### B. Related work

Detection, identification and adaption-Methods [7], such as Receiver Autonomous Integrity Monitoring (RAIM), require deep sensor understanding and redundant data. Agogino et al. [2] compare squared Mahalanobis distance with a chi-squared threshold as a consistency check, also known as NIS-test (Normalized Innovation Squared). An algorithm for combining multiple models is presented in [1]. However, correlations are neglected. Furthermore, the Kalman innovation vector has to be known, hence black-box solutions can not be checked. An algorithm for switching between vehicle dynamic models, depending on the driving situation, is presented in [12]. Chilian et al. [4] present a robust pose estimator, using visual odometry, IMU (Inertial Measurement Unit) and leg odometer to estimate the movement and pose of a 6-leg crawler. Erroneous estimates from visual odometry are reweighed using the error state vector. Suenderhauf [11] presents an approach for robust SLAM (Simultaneous Localization and Mapping) using pose graphs with switchable variables to exclude outliers during optimization.

Our approach is superior to the ones above, as we incorporate and estimate covariance / correlation matrices, reject outliers and fuse odometer estimates without the necessity of positioning sensors or large computational burden. We are furthermore able to cope with limited correlated odometer estimates. Our approach is generic and does not postulate deep sensor understanding or availability of raw sensor data such as RAIM. Additionally, we present top-to-bottom calibration of dynamic sensors.

## II. Vehicle setup

Our test vehicle is an Opel Insignia displayed in Fig. 2. We use the vehicles in-series sensors, such as the steering angle sensor (SAS), four wheel speed sensors (WSS), a 2D acceleration unit and a yaw rate gyroscope. The in-series sensors are accessible via CAN (Controller Area Network). Additionally, two gray-scale cameras[2] have been installed on the front and rear window, as shown in Fig. 2. A low-cost GNSS receiver[3] is mounted on the roof. As ground truth an INS (Inertial Navigation System) unit combined with a DGPS[4] has been installed. Latter has a high-precision 3D accelerometer and gyroscope as well as a DGPS sensor with a receiver for correction data via RTCM (Radio Technical Commission for Maritime Services) protocol. This sensor is solely used as ground truth for validation.

Additionally, two high-precision digital maps have been recorded beforehand. The first one covers 140km of highway roads around Frankfurt, Germany, including the roads between Frankfurter, Bad Homburger and Offenbacher Kreuz (A3, A5, A661). The second map covers the highway test track of the Opel testing ground in Dudenhofen, Germany.

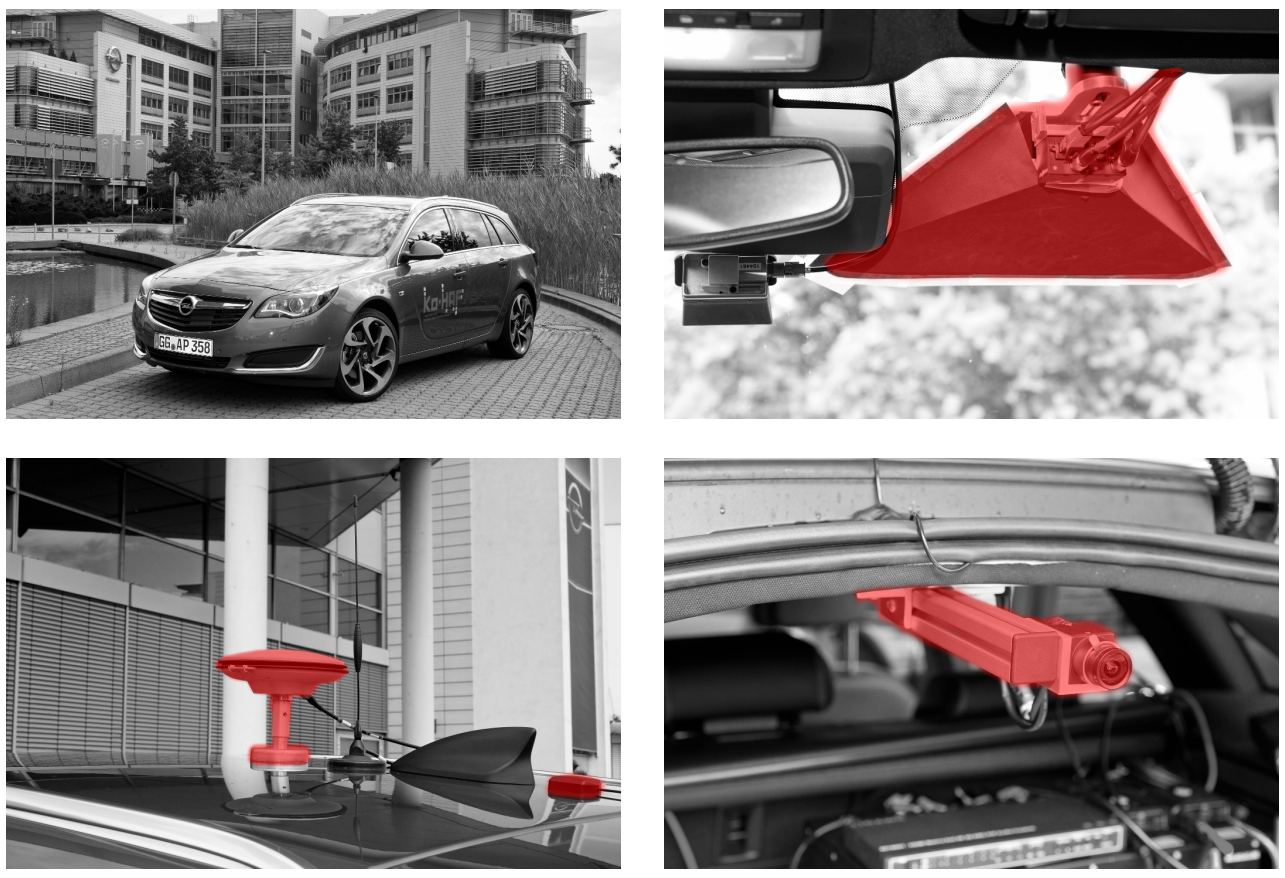

Fig. 2. **Experimental setup.** Top left: Test vehicle. Top right: Front camera setup. Bottom left: DGPS antennas. Bottom right: Lane Detection.

## III. Calibration

Sensor calibration algorithms include estimation of time delays, sensor alignment (commonly referred to as rigid transformation), offset / scaling and covariances / correlations.

### A. *Time delays*

Significant time delays of sensor measurements typically occur with increased computing time due to necessary preprocessing. GNSS pseudo- / deltarange calculus or image processing are the most prominent ones. When fusing sensor data of multiple sensors these delays need to be considered. To compute GNSS and camera time delays we cross-correlate (time domain) delta orientation of GNSS / lane relative position and visual odometry with mean IMU yaw rate in respecting time intervals. The vehicle's IMU yaw rate, provided over CAN, has negligible time delay of less than 2ms. Peaks in the resulting correlation graph indicate time delays [3].

[2]Point Grey, FL3-GE-28S4M-C

[3]u-Blox, EVK-M8T

[4]Genesys, ADMA Entry Level

### B. *Sensor Alignment*

To compute the translation between positioning sensors and acceleration sensors, we derive the position signal two times:

$$v = \frac{d}{dt}(r_0 + r_{rel}) = v_0 + \omega \times r_{rel} + v_{rel} \tag{1}$$

$$\begin{aligned} a &= \frac{d}{dt} v_{rel} \\ &= a_0 + \dot{\omega} \times r_{rel} + \omega \times (\omega \times r_{rel}) + 2\omega \times v_{rel} + a_{rel} \end{aligned} \tag{2}$$

where $r_0$ fis the position sensor location, $r_{rel}$ is the relative position of the acceleration sensor in vehicle coordinates and $\omega$ is the angular rate. Since $v_{rel}$ and $a_{rel}$ are 0 (rigid transformation), relative and Coriolis accelerations cancel out. Relative position $r_{rel}$ of GNSS receiver will be computed by optimizing the difference between differentiated doppler-speed vector and IMU acceleration signals (see Fig. 3). The in-series IMU is assumed to have negligible time delay / jitter and is mounted in a well known position. The relative vector can only be observed in dynamic scenarios when high yaw rates yield Euler and centrifugal accelerations. For the estimation of the front and rear camera intrinsic and extrinsic parameters, we use an approach presented by Strauss et al. [10]. This approach can be used for cameras with non-overlapping views using coded checkerboard targets.

### C. *Offset / Scaling*

While vehicle dynamic sensors (IMU, WSS, SAS) are mounted in well-known positions with negligible time delay, they typically suffer from offset and scaling errors. Offset and scaling errors can be directly estimated using Recursive Linear Regression. Fig. 7 shows GNSS delta orientations and integrated IMU yaw rate between those delta orientations. Steering angle ratio and offset is calibrated using recursive linear regression by comparing kinematic single-track model [9] with small angle approximation $\theta$ and calibrated yaw rate:

$$\dot{\psi} = \frac{1}{l_f + l_r} v\delta \tag{3}$$

where $l_f$ and $l_r$ denote the distance between the COG (center of gravity) and the front and rear axis of the vehicle. Each wheel of our test vehicle is equipped with a cyclic counter measuring the amount of rotations. A total amount of 48 ticks is equivalent to a full rotation. However, to compute velocity or traveled distance the wheel diameter has to be estimated precisely. Wheel diameters are usually estimated using positioning sensors like GNSS. Though the estimated traveled path between position measurements is larger than the true traveled path as position uncertainty increases (see Fig. 5). The mean path error between two probability density functions (PDF) can be computed by integrating Eq. 4. Where $p_i$ and $p_j$ represent 3D points (x,y,z) of a PDF $f_1$ and $f_2$ respectively. However, since most PDFs have no indefinite integral (eg. Gaussian distribution) a numerical integration would be required, which is numerically expensive as each point $p_i$ and $p_j$ has three dimensions (x,y,z) that need to be integrated in the inverval $(-\infty, \infty)$.

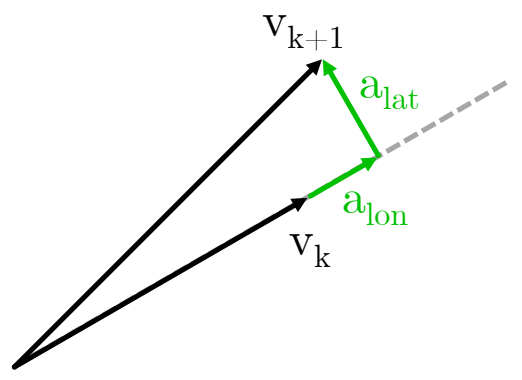

Fig. 3. **Numerical differentiation of velocity**

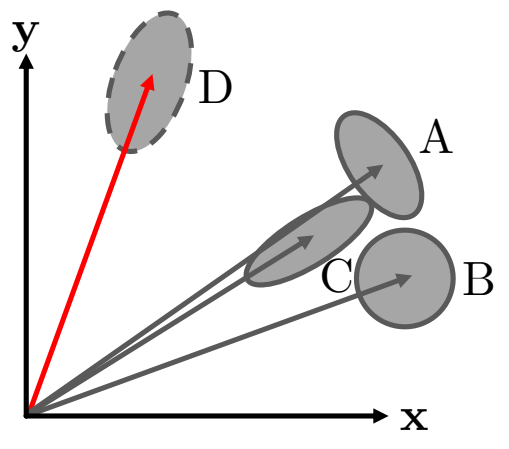

Fig. 4. **Outlier detection**

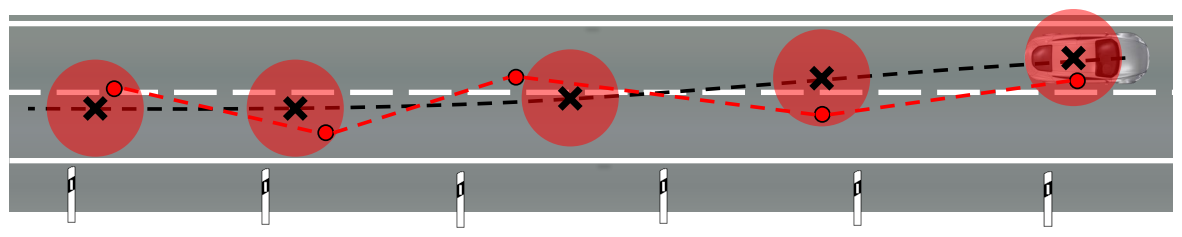

Fig. 5. **Absolute path versus measured path**. The estimated path (red) between sensor measurements is bigger than the actual path (black). The sensor uncertainty is represented by the big, red circles.

$$\iiint_{p_i} \iiint_{p_j} f_1(p_i) f_2(p_j) ||p_i - p_j|| dp_i dp_j \tag{4}$$

Without loss of generality we define x to be the direction between of movement. The following assumptions simplify the posed problem:

1) The positioning systems PDF is a 3D multivariate normal distribution with zero mean and diagonal covariance matrix $\Sigma$.
2) The diagonal elements of $\Sigma$ are equal for all three dimensions ($\sigma_{xx} = \sigma_{yy} = \sigma_{zz}$).

The first assumption is usually valid as covariance matrices are similarly aligned between two measurements. The second assumption may usually be violated. For example GNSS measurements tend to have higher uncertainties in altitude estimation than in horizontal estimation. We will investigate violation of the second assumption later. We derive the formula for the distance between two spheres using spherical coordinates:

$$d_{ss}(d, R_1, R_2) = d + \frac{R_1^2 + R_2^2}{3d} \qquad \forall d > R_1 + R_2$$

where $d$ is the distance between the spheres and $R_1$ and $R_2$ are the sphere radii. Using the Maxwell Boltzmann distribution:

$$P_R(r) = \iint_D f(p_i) dxdydz = .. = \frac{2}{\sqrt{2\pi\sigma^3}} r^2 \exp(-\frac{1}{2\sigma^2} r^2)$$

the sphere distance formula and the definite integral of the Gaussian function, we can algebraically solve:

$$\begin{aligned} d_{err} &= \int_{r_1=0}^{\infty} \int_{r_2=0}^{\infty} P_R(r_1) P_R(r_2) d_{ss}(d, r_1, r_2) \\ &= \int_{r_1=0}^{\infty} \int_{r_2=0}^{\infty} [3d^2 r_1^2 r_2^2 + r_1^4 r_2^2 + r_1^2 r_2^4] ... \\ &... \exp(-\frac{r_1^2}{2\sigma_1^2}) \exp(-\frac{r_2^2}{2\sigma_2^2}) \\ &= d + \frac{\sigma_1^2 + \sigma_2^2}{d} \end{aligned}$$

which can be solved for d. To account for varying uncertainties along the 3D space, we compute a mean covariance from perpendicular covariance $\sigma_{yy}$ and $\sigma_{zz}$:

$$\sigma_{est}^2 = \frac{\sigma_{yy}^2 + \sigma_{zz}^2}{2} \tag{5}$$

It is important to motivate that the influence of sensor uncertainty can be decreased when a filter (e.g. Kalman filter) is applied. However, the results still has uncertainties, although smaller, that distort the absolute traveled path estimation.

### D. *Covariance / Correlation Estimation*

Covariance matrices between two samples $X_1$ and $X_2$ can be computed using the following formula:

$$\Sigma_{1,2} = E[(X_1 - \mu_1)(X_2 - \mu_2)^T]$$

We assume that all of our odometers have multivariate normal distribution with mean $\mu_{true}$, thus $\mu_1 = \mu_2 = \mu_{true}$. To compute the covariance of and correlation between odometry estimates (e.q. visual odometry, single-track model, WSS+IMU, ..) the mean $\mu_{true}$ needs to be known, but can not be estimated using the sample mean as it is different for each measurement. The idea is to subtract odometry estimates, which results in:

$$\begin{aligned} &X_1 \sim \mathcal{N}(\mu_1, \Sigma_1) && X_2 \sim \mathcal{N}(\mu_2, \Sigma_2) \\ &X_{12} = X_1 - X_2 && \mu_1 = \mu_2 \end{aligned}$$

$$\begin{aligned} &X_{12} \sim \mathcal{N}(0, \Sigma_1 + \Sigma_2) && \text{if } X_1, X_2 \quad \text{uncorrelated} \\ &X_{12} \sim \mathcal{N}(0, \Sigma_2 - 2\Sigma_{12} + \Sigma_1) && \text{if } X_1, X_2 \quad \text{correlated} \end{aligned}$$

For $k$ uncorrelated odometers there are $k(k-1)/2$ possible combinations. Eq.6 shows the $k=3$ and the general case, if all odometry estimates are uncorrelated.

$$\begin{bmatrix}\tilde{\Sigma}_{1-2}\\ \tilde{\Sigma}_{1-3}\\ \tilde{\Sigma}_{2-3}\end{bmatrix} = \begin{bmatrix} I_j & I_j & 0\\ I_j & 0 & I_j\\ 0 & I_j & I_j\end{bmatrix}\begin{bmatrix}\tilde{\Sigma}_1\\ \tilde{\Sigma}_2\\ \tilde{\Sigma}_3\end{bmatrix} \tag{6}$$

$$\begin{bmatrix}\tilde{\Sigma}_{1-2}\\ \vdots\\ \tilde{\Sigma}_{1-k}\\ \tilde{\Sigma}_{2-3}\\ \vdots\\ \tilde{\Sigma}_{(k-1)-k}\end{bmatrix} = \begin{bmatrix} I_j & I_j & 0 & 0 & \cdots\\ I_j & 0 & I_j & 0 & \cdots\\ I_j & 0 & 0 & I_j & \cdots\\ \vdots & \vdots & \vdots & \vdots & \ddots\\ 0 & I_j & I_j & 0 & \cdots\\ 0 & I_j & 0 & I_j & \cdots\\ \vdots & \vdots & \vdots & \vdots & \ddots\end{bmatrix}\begin{bmatrix}\tilde{\Sigma}_1\\ \vdots\\ \tilde{\Sigma}_k\end{bmatrix} \tag{7}$$

where $\tilde{\Sigma}$ is the column vector of upper triangular matrix of $\Sigma \in \mathbb{R}^{nxn}$ with size $j = (n(n-1)/2)$. The matrices can then be computed solving the least squares problem:

$$b = Ax \qquad\qquad x = (A^TA)^{-1}A^Tb$$

If samples are correlated the number of unknowns increases by $k_{corr}$. For each correlated sample set $xy$ we strike out the respective row in Eq. 7 and solve for the correlation matrix later:

$$\Sigma_{xy} = \frac{1}{2}(\Sigma_x + \Sigma_y - \Sigma_{x-y}) \tag{8}$$

Note that if too many samples are correlated the problem can not be solved. The whole calibration process is depicted in Fig. 6.

## IV. Fusion and Outlier Rejection

With the calibrated sensor setup and known covariances/correlations, we detect erroneous odometry outliers using state-of-the-art NIS-test. If a sensor has larger Mahalanobis distance than expected (using $\chi^2$-table), then the estimate is rejected. Fig. 4 shows the exclusion of transformation D from a subset of four transformations A-D. If less than three transformations are available, past transformations can be used for consistency check, after they have been updated to the current time. This requires an accurate propagation algorithm for transformations, which we implemented using our calibrated IMU:

$$dx_{k+1} = dx_k + n \cdot a_{lon} + n_{\perp}a_{lat} \tag{9}$$

where $n$ is the direction of $dx_k$ The resulting transformation between environmental perceptions will then be computed by combining all valid transformations iteratively (using Eq. 10). Note that correlated estimates are fused using their uncorrelated covariance matrices and adding their cross-correlation matrix to the uncertainty of the resulting estimate.

$$\begin{aligned} K_k &= P_{aa}H^T(HP_{aa}H^T + P_{bb})^{-1}\\ c &= a + K_k(b - Ha)\\ P_{cc} &= P_{aa} - K_kHP_{aa}\end{aligned} \tag{10}$$

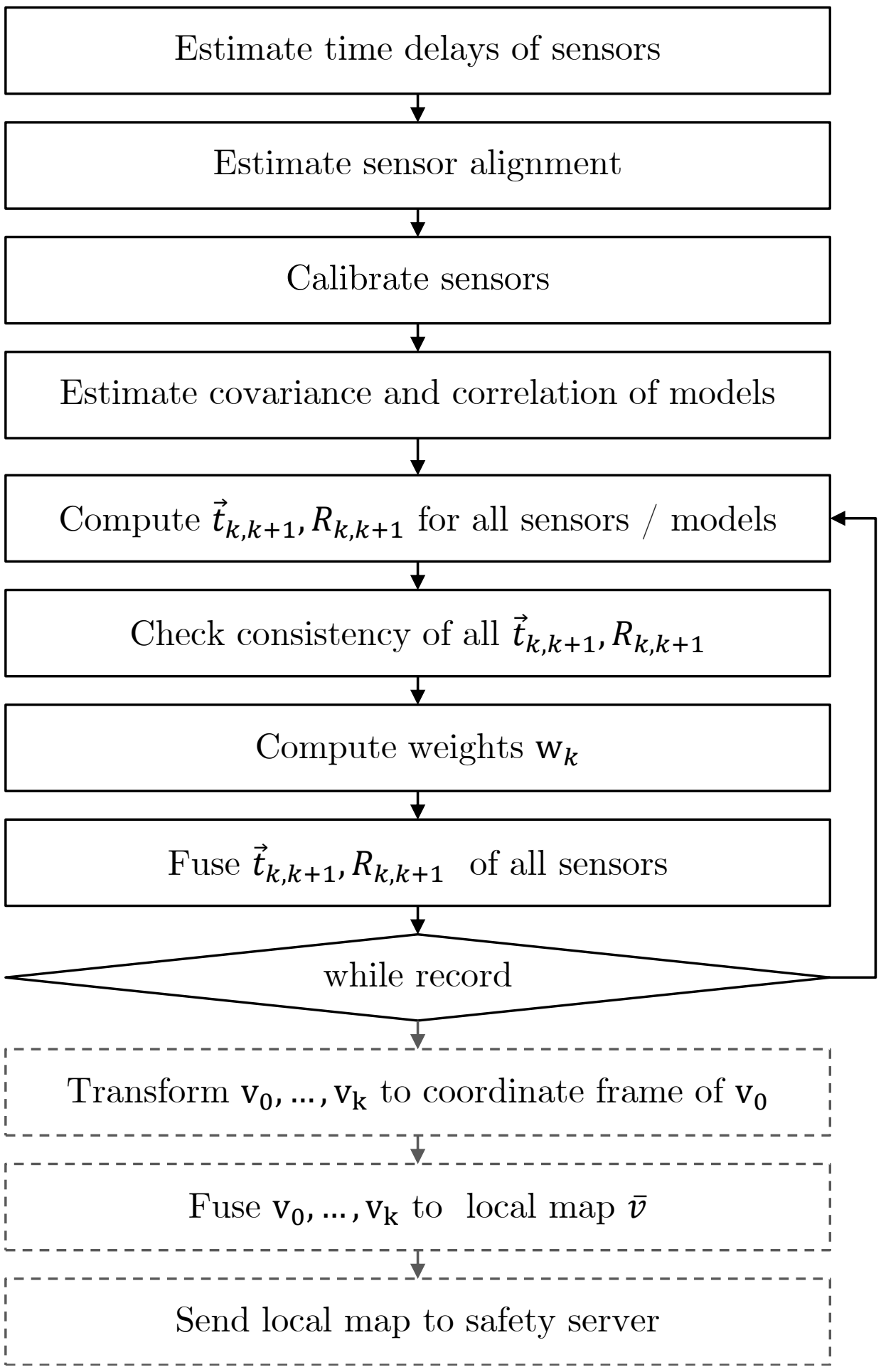

Fig. 6. **Flow chart.** Calibration, covariance estimation, outlier rejection and fusion.

where $a$ and $b$ are the odometer estimated with $P_{aa}$ and $P_{bb}$ covariance matrices. $c$ is the resulting estimate with covariance matrix $P_{cc}$. The formula has been derived by Kalman [6] and is often referred to as the Kalman estimation or Kalman update.

## V. Experimental results

Fig. 7 shows the GNSS yaw rate estimation and the yaw rate measurement from the vehicle IMU. Numerical differentiation amplifies high frequencies due to the fact that it is similar to multiplication of absolute frequency in Laplace scope. Thus, Fig. 7 (top) shows high noise in GNSS yaw rate estimation. The effect can be, for qualitative analysis, suppressed with a mean (or low-pass) filter shown in Fig. 7 (bottom). Fig. 8 shows the correlation result of the GNSS yaw rate estimation with the IMU yaw rate. Peaks in the resulting correlation graph indicate time delays. For the depicted data set of 200 s a time delay of the GNSS sensor of 130 ms has been found.

To show the performance of our improved path estimation algorithm for wheel diameter estimation, we generated 5000 samples along the x-axis with distance $d$ to simulate a straight movement. Additionally, each sample has random 3D normal

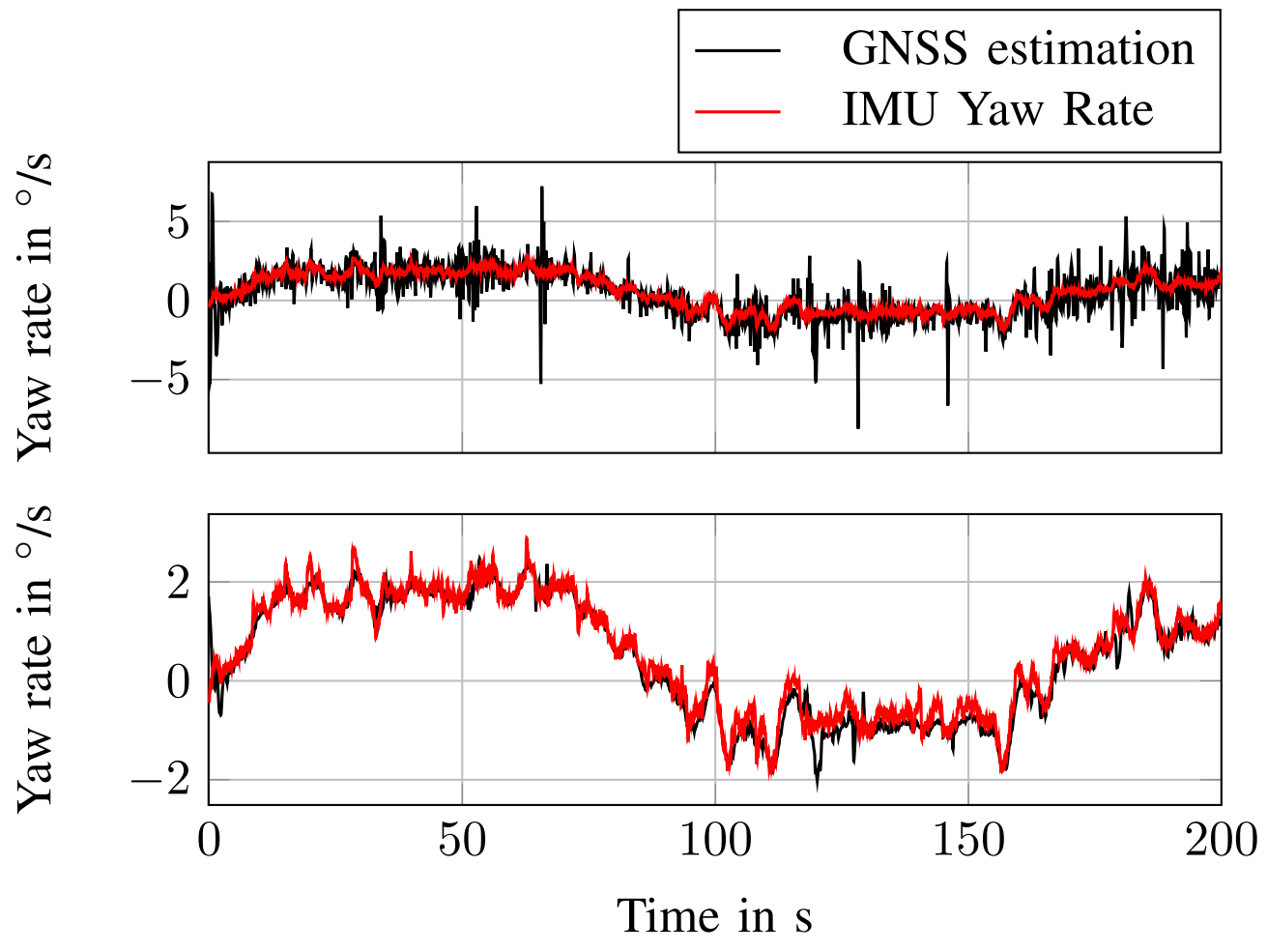

Fig. 7. **Derived GNSS orientation**. The top graph shows the raw GNSS yaw rate estimation. The bottom graph shows the filtered GNSS yaw rate estimation (sliding window filter, size=11).

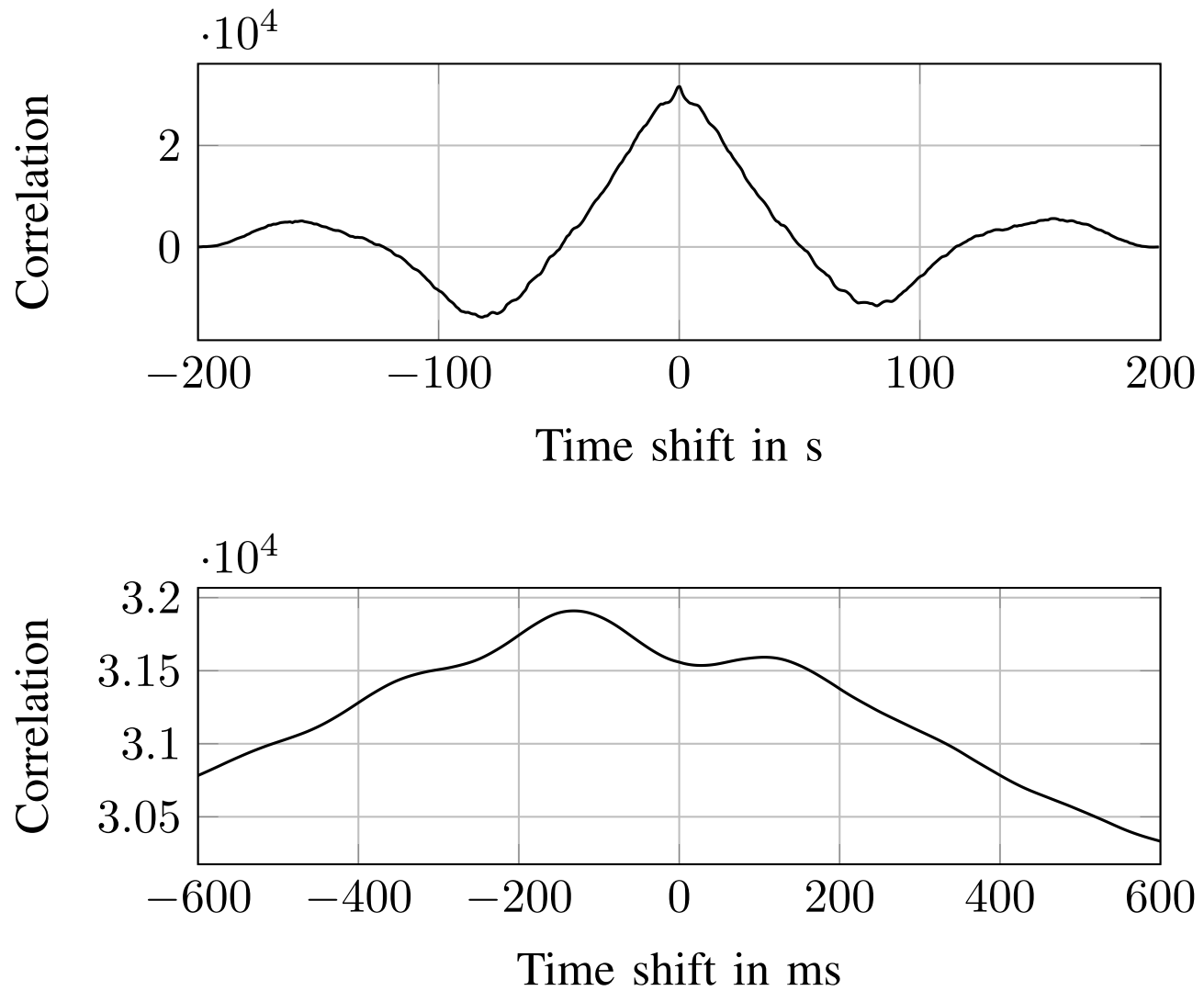

Fig. 8. **GNSS time delay estimation.** The maximal value of the correlation graph has been found to be at 130 ms.

distribution noise to simulate sensor uncertainty. Fig. 9 (top) shows the path error for $\sigma_{est} = \sigma_{xx} = \sigma_{yy} = \sigma_{zz} = 1$ with increasing $d$ along the x-axis. Fig. 9 (bottom) shows the path error for $\sigma_{xx} = \sigma_{yy} = 1/3 * \sigma_{zz} = 1$ which violates our assumption of equal noise in all dimensions. Hence a mean covariance $\sigma_{est}$ as stated in Eq. 5 is computed. As the ratio of $d$ and $\sigma_{est}$ increases, the error decreases exponentially, solely because the sensor noise has negligible influence when the distance increases. However, omitting GNSS measurement to increase the ratio, results in inferior path estimations in non-straight movements. Fig. 9 shows that the corrected path is approximately one magnitude smaller than the path estimation

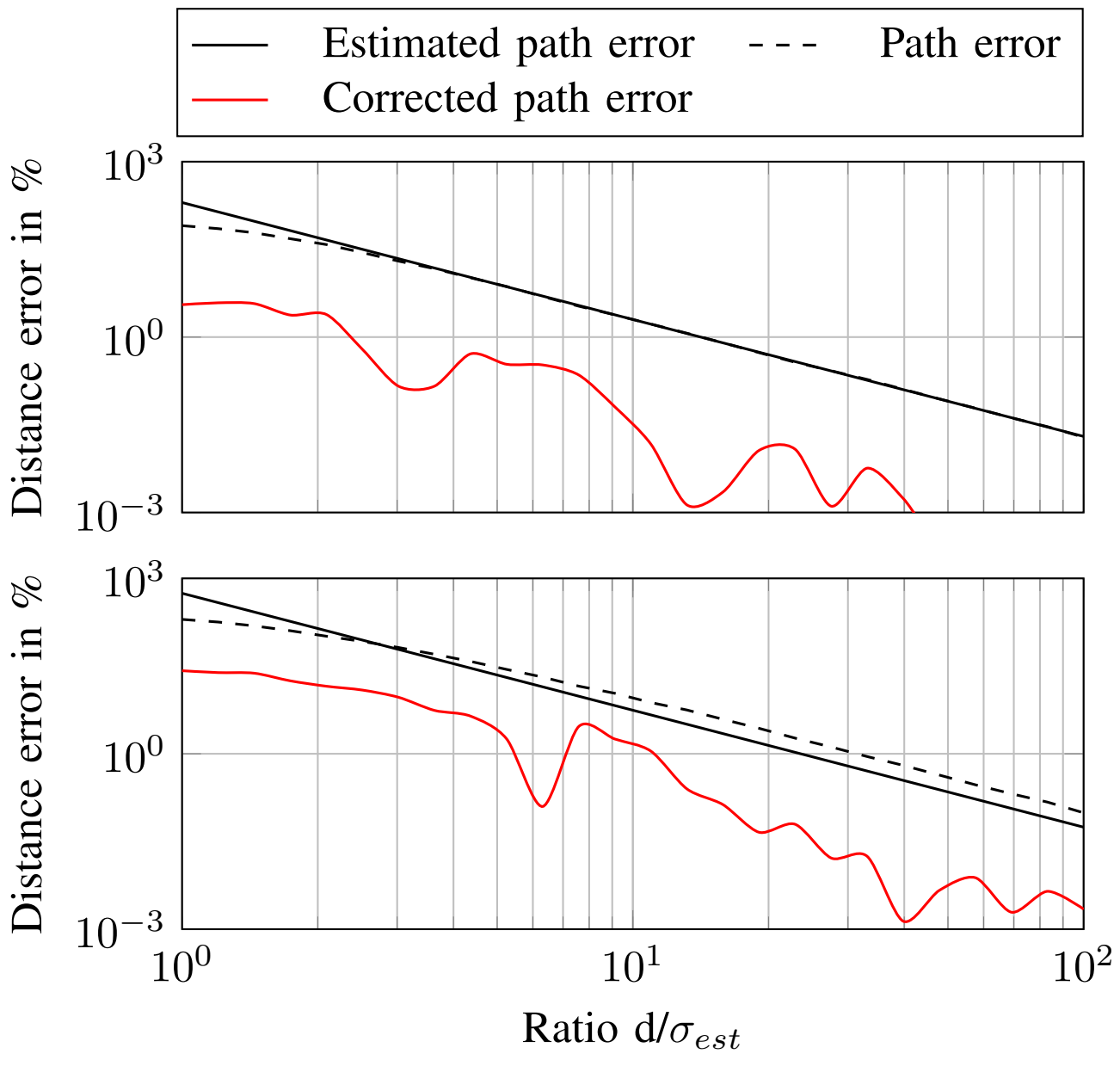

Fig. 9. **Wheel diameter calibration**. 5000 3D normally distributed samples per data point with increasing distance d along x-axis. Top: $\sigma_{xx} = \sigma_{yy} = \sigma_{zz} = 1$. Bottom: $\sigma_{xx} = \sigma_{yy} = 1/3 * \sigma_{zz} = 1$

when neglecting the sensor noise.

After the GNSS time delay has been computed, the orientation estimates and delta velocities (see Fig. 3) are used to estimate offset and scale of the vehicle IMU via recursive least squares. Fig. 10 shows the IMU yaw rate offset and scale estimation using the GNSS sensor (red) compared to the estimation using the high-precision accelerometer and gyroscope of the ADMA sensor (black, dashed).

| Yaw rate | Offset | Scale |
|---|---|---|
| ADMA | -0.1626 deg/s | 1.0576 |
| GNSS | -0.1542 deg/s | 1.0495 |

To validate our approach of computing the covariance and correlation matrices (Eq. 7) 10, 100, 1000 and 10000 samples have been drawn from normal distributions, displayed in Fig. 11 as ellipsoids, with a normally distributed mean that simulates random, unknown movement. Dark colors represent the covariance matrices where the samples are drawn from and bright colors represent the estimation of the covariance matrix. As sample size increases the estimation becomes more accurate. Green and red samples are correlated with the black ellipsoid. Fig. 12 shows the online covariance estimation of our odometers under the ROS framework [8].

Fig. 13 shows our robust odometry (black) among other odometer estimates. Thin black lines show the digital map data of the Offenbacher Kreuz. The initial point for all odometers is displayed as green dot. Fig. 14 shows the detected lanes (red) from multiple camera images. Translation and rotation between images has been computed using our robust odometry. The absolute position is estimated using the GNSS sensor. The digital map is displayed as black lines. The lanes can now be assigned and fused.

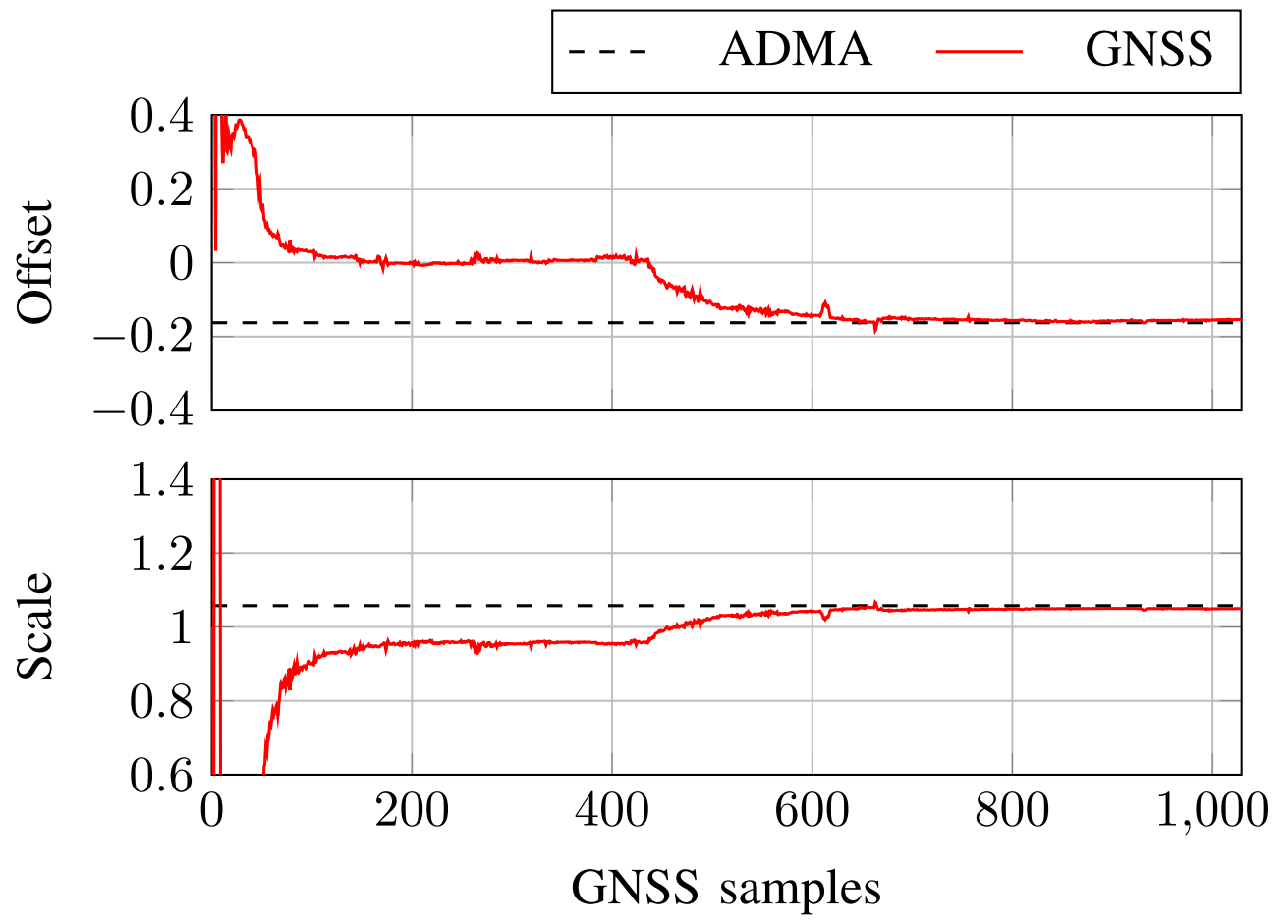

Fig. 10. **Yaw rate offset and scale estimation**. The estimation using the GNSS measurements (red) converges to the estimation using the high-precision gyroscope of the ADMA sensor.

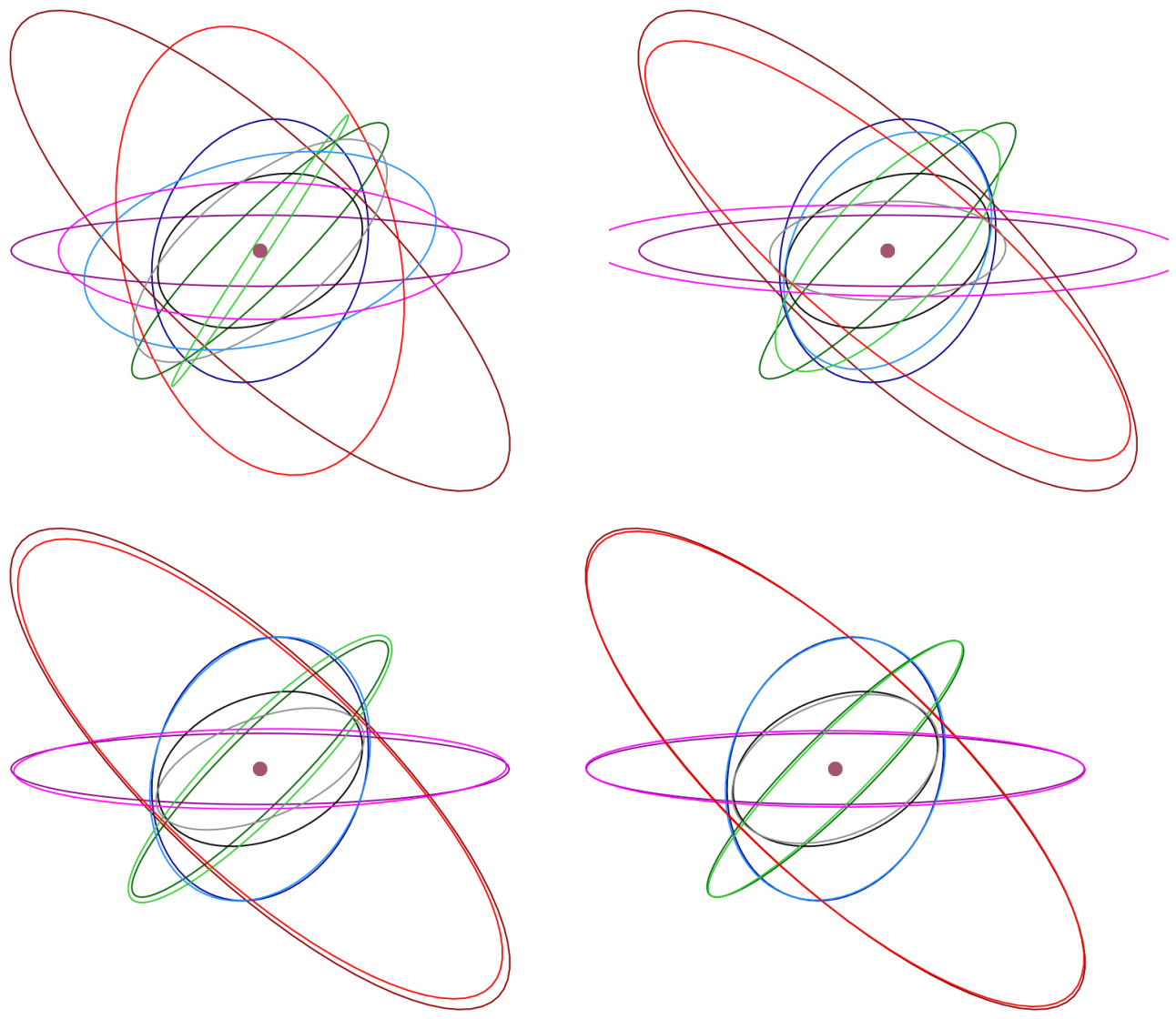

Fig. 11. **Covariance estimation**. Top left: 10 samples. Top right: 100 samples. Bottom left: 1000 samples. Bottom right: 10000 samples. Bright colored ellipsoids represent the estimated covariance matrices. Gray ellipsoid represents the correlation matrix between red and green samples.

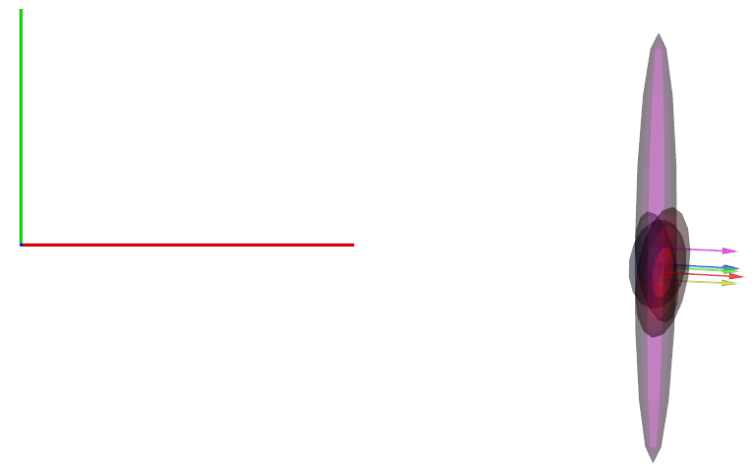

Fig. 12. **Online covariance estimation of odometers**. Each ellipsoid represents the estimated uncertainty of each odometer.

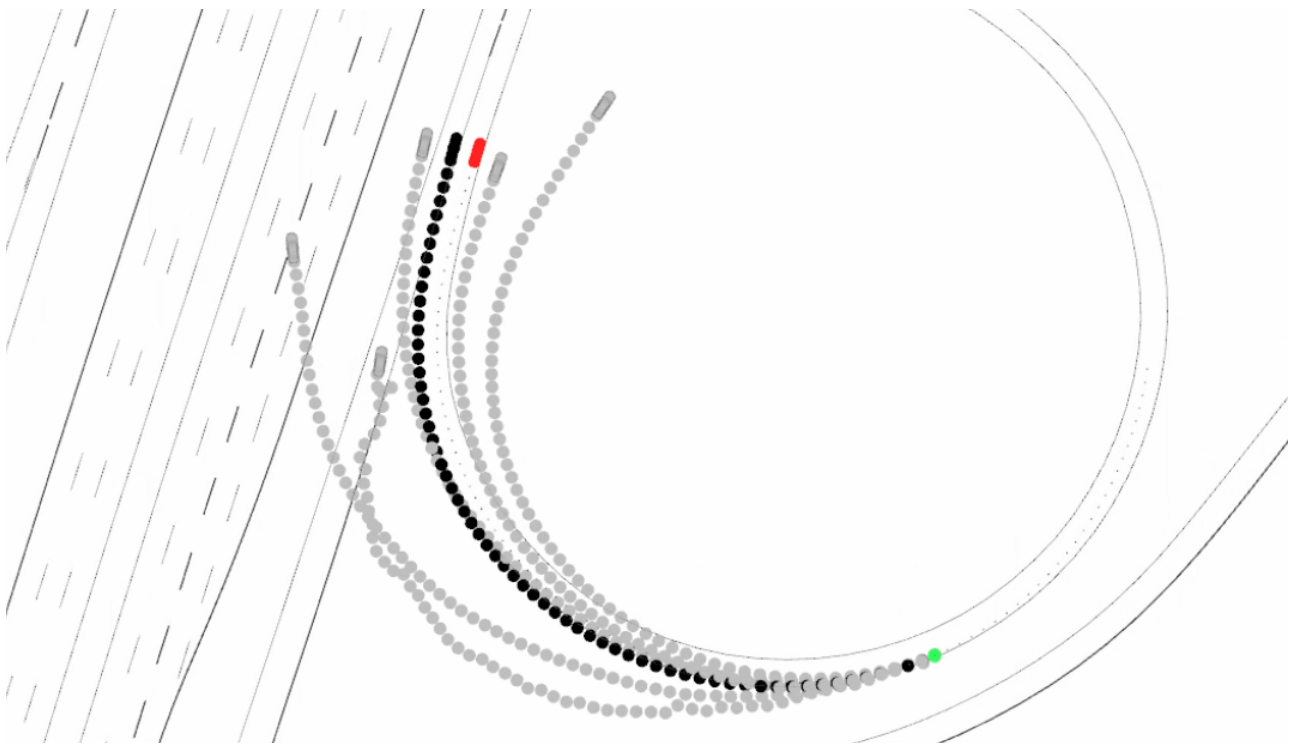

Fig. 13. **Robust odometry**. Gray lines show odometer estimates. Red dot represents the current vehicle position. Black line is the robust odometer estimation. Green dot is the inital starting point of all odometers.

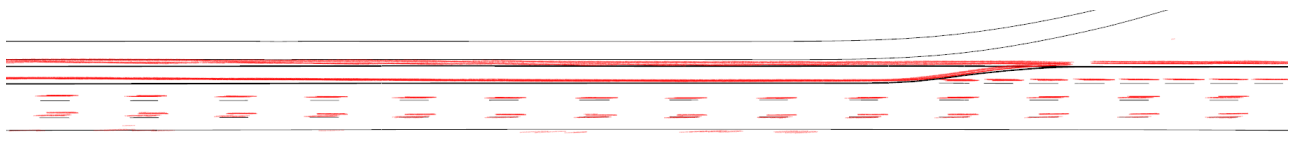

Fig. 14. **Map data with robust odometry**. Red lines represent detected lane markings. Black lines represent the previously recorded map.

## VI. Conclusions and Future Work

In this paper we presented a top-to-bottom calibration cycle with covariance / correlation matrix estimation. A new approach to estimate wheel diameters, using a corrected total path estimation between normally distributed samples, has been presented. Furthermore, a generic approach to estimate covariance / correlation matrices between various odometers has been derived. We use state-of-the-art NIS-test to check odometers for consistency and to exclude erroneous measurements. The odometer estimates with their known uncertainties can then be fused using the Kalman estimation equations.
Our approach for robust odometry can furthermore be used to check consistency of positioning signals (see Fig. 15). With known covariances of the odometers, the covariance of the positioning sensor can easily be computed. Additionally, prior position estimates can be used to check new positioning signals at time $t_k$, as displayed in Fig. 15. The robust odometer can be used to project prior estimates to time $t_k$.

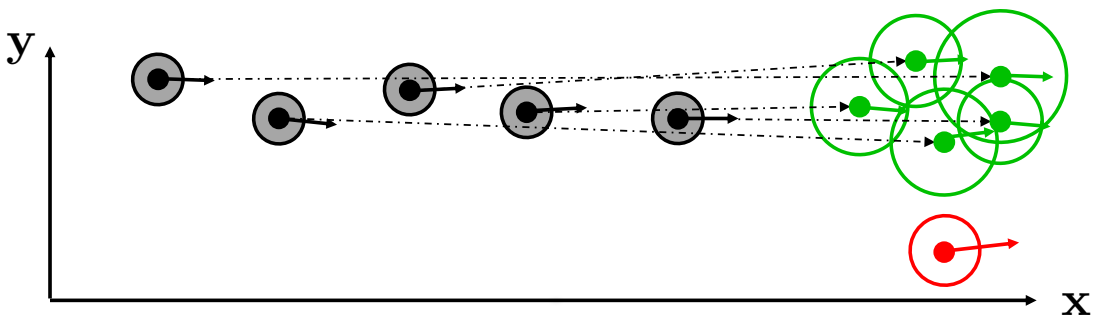

Fig. 15. **Outlier rejection of position estimates**. Prior position estimates (black) are updated / propagated using the robust odometer (green) and compared to the latest position estimate (red).

## VII. ACKNOWLEDGMENTS

We thank the Opel Concept & Teardown Workshop for their indispensable support and help with the Opel Insignia, including but not limited to Michael Dahlke and Roman Rudek. We furthermore thank the Opel Automobile GmbH for their support. Last but not least we would like to thank the Ko-HAF team for all the enrichments and thoroughly good discussions.